# Plain language adaptations of biomedical text using LLMs: Comparision of evaluation metrics

Primoz KOCBEK [a,b,1] Leon KOPITAR [a], and Gregor STIGLIC [a,c]
[a] *University of Maribor, Faculty of Health Sciences, Maribor, Slovenia*
[b] *University of Ljubljana, Faculty of Medicine, Ljubljana, Slovenia*
[c] *University of Edinburgh, Usher Institute, Edinburgh, UK*
ORCiD ID: Primoz KOCBEK https://orcid.org/0000-0002-9064-5085
Leon KOPITAR https://orcid.org/0000-0002-6647-9988
Gregor STIGLIC https://orcid.org/0000-0002-0183-8679

**Abstract.** This study investigated the application of Large Language Models (LLMs) for simplifying biomedical texts to enhance health literacy. Using a public dataset, which included plain language adaptations of biomedical abstracts, we developed and evaluated several approaches, specifically a baseline approach using a prompt template, a two AI agent approach, and a fine-tuning approach. We selected OpenAI gpt-4o and gpt-4o mini models as baselines for further research. We evaluated our approaches with quantitative metrics, such as Flesch-Kincaid grade level, SMOG Index, SARI, and BERTScore, G-Eval, as well as with qualitative metric, more precisely 5-point Likert scales for simplicity, accuracy, completeness, brevity. Results showed a superior performance of gpt-4o-mini and an underperformance of FT approaches. G-Eval, a LLM based quantitative metric, showed promising results, ranking the approaches similarly as the qualitative metric.

## 1. Introduction

Individual Health literacy, defined as the degree to which individuals can find, understand, and use information and services to inform health-related decisions and actions for themselves and others [1], is seen as a key component in public health policy and set as a priority by research agencies around the world. For example, in the United States the Agency for Healthcare Research and Quality (AHRQ) released a call titled Research on Improving Organizational Health Literacy to Prevent and Manage Chronic Disease [2]. There are also initiatives, such as European Health Literacy Survey (HLS-EU) [3], where results showed significant disparities in health literacy levels across member states, such as that at least 1 in 10 (12%) respondents showed insufficient health literacy and almost 1 in 2 (47%) had limited (insufficient or problematic) health literacy.

Traditional health literacy interventions, consisting of lectures, passive lessons, one-way delivery of information, distribution of pamphlets and leaflets, and health-education sessions with visual aids, have shown no clear effectiveness on people with low literacy, sustainability, and scalability. We believe that these methods lack the precision and individualization required for optimal results in improving health literacy. This might be bridged Large Language Models (LLMs), which can generate human-like text with remarkable accuracy for multiple downstream tasks [3].

---

[1] Corresponding Author: Primoz KOCBEK, 2000 Maribor, +386 2 300 47 13, primoz.kocbek@um.si, University of Maribor, Faculty of Health Sciences, Maribor, Slovenia

In this preliminary work a public dataset of biomedical abstracts with included plain language adaptations was used to develop LLM approaches and be evaluated by both qualitative (human) as well as quantitative (automatic) metrics for plain language adaptations at a specific level of readability (<K8), i.e. students 13–14 years old, as recommended by National Institutes of Health (NIH) for written health materials [4].

**2. Methods**

*2.1. Data preparation*

The public PLABA dataset [5] was used as a reference dataset and consists of 750 manually adapted abstracts that are both document- and sentence-aligned. The abstracts have been retrieved to answer consumer questions asked on MedlinePlus [5]. More precisely, there are 75 questions, each containing 10 questions and we split the data 80% for training (733 samples) and 20% for validation (184 samples) with respect to '*pmid*' document level identifier since the abstracts could have more than one adaptation.

The adaptations were created using guidelines that assume that the average literacy level is lower than grade 8 (<K8) as recommended by NIH for written health materials [6]. The plain language adaptations were sentence level and included the following methods: term substitution of jargon with a common alternative, term explanation if no substitution exists, term generalization if significance is not lost for general audience, term removal if explanation is too technical and not relevant to understanding [6].

*2.2. Models*

Since a public dataset was used, we decided to use a GDPR compliant version of OpenAI models via api, more specifically we used gpt-4o-mini, model version `gpt-4o-mini-2024-07-18` and gpt-4o, model version `gpt-4o-2024-08-06` for our approaches.

We compared three approaches for each model, the first being a simple prompt template with prompt engineering, the second being an expansion of the first one by using two AI agents iteratively where they improved the output and lastly a fine-tuned (FT) models. In the two AI agent approach we created a discussion, called a thread, for each input, where the first agent created an adaptation using the baseline prompt and the second AI agent in the persona of a "smart 13–14-year-old student" asked clarification questions, on the basis of which the first AI agent modified the adaptation. In the FT approach `gpt-4o` and `gpt-4o-mini` were FT on the training data via the OpenAI api dashboard. We used the following hyperparameters: epochs 3, batch size 1, LR multiplier 2, random seed 741667963. The results from FT were as follows: `gpt-4o` training loss 1.099, full validation loss 0.8336 and `gpt-4o-mini` training loss 1.0489, full validation loss 0.967.

We used around 1,9 M trained tokens for each FT run, currently that is $50 for `gpt-4o` and $6 for `gpt-4o-mini`. We estimated that we used 1 M tokens for each baseline approach and 2M for the two-agent approach, which considering prices of $10 for `gpt-4o`and $0.6 for `gpt-4o-mini` per 1M tokens, totals of around $100 [7].

We produced six end-to-end adaptations, in. the rest of the paper they are referred as `gpt-4o_baseline, gpt-4o_two_agents, gpt-4o-ft, gpt-4o-mini_baseline, gpt-4o-mini_two_agents, gpt-4o-mini-ft`.

*2.3. Evaluation metrics*

The problem in using LLMs for a nonstandard tasks is finding and using appropriate evaluation metrics. In many cases the gold standard is still post-hoc human evaluation of domain experts, such as using 5-, 6- or 10- point Likert scales for concepts like sentence simplicity, term simplicity, term accuracy, fluency, completeness, faithfulness [5].

We used multiple quantitative metrics: two were common readability scores, Flesch-Kincaid (FK) grade level [8] and SMOG [9], both are US grade level based, i.e. it should be lower than 8 (<K8) or students 13–14 years old [4], the second was SMOG Index, however the index for fewer than 30 sentences is statistically invalid since the formula was normed on 30-sentence samples and our abstract were the length 10 sentences [9], we therefore primarily use FK grade level results. Additionally, we used a general simplification metric SARI [10], a semantic similarity metric BERTScore [11], and a LLM based metric G-Eval [12], which is a framework that uses chain-of-thoughts (CoT) LLMs to evaluate the outputs based on defined criteria. In our case we used the average of the 4 categories used for qualitative evaluation as described below.

For qualitative evaluation we used a 5-point Likert scale for simplicity, i.e. outputs should be easy to understand, accuracy, i.e. outputs should contain the accurate information, completeness, i.e. outputs should seek to minimize information lost from the original text and brevity, i.e. outputs should be concise [6]. We adapted the categories from the Plain language adaptation of biomedical abstracts (PLABA) 2024 end-to-end abstract adaptation competition [6].

## 3. Results

The quantitative metrics were evaluated on the validation data, since we used the training data in the FT approach. We observed the FK grade level was at college level with an average of 13.67 (SD=3.29), the plain language adaptation (ground truth) drops to 12th grade level with an average of 11.64 (SD=2.43), which was much higher than expected. The FT approaches got the closest to the grade levels of plain adaptations, the `baseline` and the `two_agents` approaches got closer to the 8th grade level line (Figure 1).

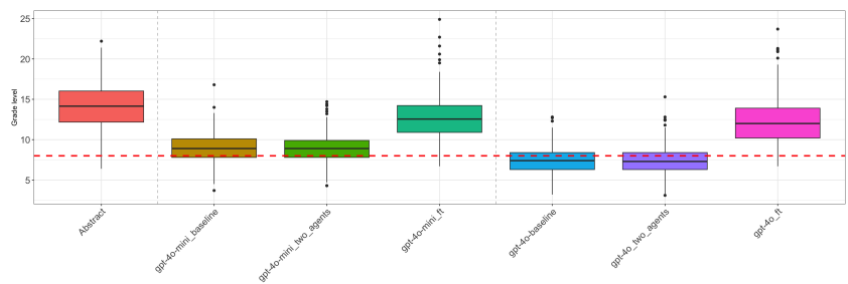

**Figure 1.** Visual representation of FK grade level for Abstract, gpt-4o and gpt-4o-mini approaches and ground truth on the training validation set.

**Table 1.** Quantitative evaluation for FK grade level, SMOG Index, SARI and G-Eval.

| Model | FK grade level | SMOG Index | BERTScore | SARI | G-Eval* |
|---|---|---|---|---|---|
| gpt-4o-mini_baseline | 8.93 (SD=1.76) | 11.12 (SD=1.49) | 0.90 (SD=0.02) | 42.93 (SD=7.70) | 0.79 (SD=0.16) |
| gpt-4o-mini_two_agents | 8.91 (SD=1.85) | 11.11 (SD=1.52) | 0.90 (SD=0.02) | 43.17 (SD=7.41) | 0.81 (SD=0.17) |
| gpt-4o-mini_ft | 12.16 (SD=2.70) | 13.98 (SD=2.10) | 0.90 (SD=0.02) | 46.74 (SD=5.72) | 0.73 (SD=0.14) |
| gpt-4o-baseline | 7.40 (SD=1.72) | 9.54 (SD=1.67) | 0.89 (SD=0.02) | 37.64 (SD=8.49) | 0.74 (SD=0.11) |
| gpt-4o_two_agents | 7.29 (SD=1.59) | 9.60 (SD=1.63) | 0.89 (SD=0.02) | 37.50 (SD=8.20) | 0.75 (SD=0.14) |
| gpt-4o_ft | 12.20 (SD=2.76) | 13.89 (SD=2.31) | 0.90 (SD=0.02) | 45.82 (SD=6.26) | 0.74 (SD=0.13) |

*Average of simplicity, accuracy, completeness and brevity categories*

Looking at the simplification metric SARI and semantic similarity metric BERTScore we observed similar performance, where SARI showcased a somewhat better performance for `gpt-4o-ft`. G-Eval showed better performance for `gpt-4o-mini-baseline` and `gpt-4o-mini-two_agents` approaches (Table 1).

We used qualitative (human) evaluations by five healthcare experts with at least MSc level of knowledge in a the healthcare field on a sample of n=40 abstracts, where we observed that some models perform better at some categories than others, for example `gpt-4o-baseline` and `gpt-4o-two_agents` outperformed in simplicity and brevity however they perform poorly in accuracy and completeness. The best performing approaches were `gpt-4o-mini-baseline` and `gpt-4o-mini-two_agents`, both performed on average above 4 out of 5 on all evaluation categories (Table 2). We also calculated a standardized average of the 4 categories and they were similar with the quantitative evaluations.

**Table 2.** Qualitative evaluation of a sample (n=40) for simplicity, accuracy, completeness, brevity and standardized average.

| Model | Simplicity | Accuracy | Completeness | Brevity | Average (normalized) |
|---|---|---|---|---|---|
| gpt-4o-mini_baseline | 4.08 (SD=1.02) | 4.2 (SD=0.88) | 4.42 (SD=0.75) | 4.03 (SD=0.77) | 0.714 (SD=0.060) |
| gpt-4o-mini_two_agents | 4.22 (SD=0.95) | 4.25 (SD=0.87) | 4.38 (SD=0.73) | 4.08 (SD=0.89) | 0.719 (SD=0.057) |
| gpt-4o-mini_ft | 3.75 (SD=0.93) | 4.1 (SD=0.90) | 4.3 (SD=0.73) | 3.6 (SD=0.90) | 0.637 (SD=0.081) |
| gpt-4o-baseline | 4.32 (SD=0.73) | 3.88 (SD=0.61) | 3.45 (SD=0.71) | 4.28 (SD=0.82) | 0.644 (SD=0.079) |
| gpt-4o_two_agents | 4.45 (SD=0.71) | 3.62 (SD=0.90) | 3.7 (SD=0.80) | 4.25 (SD=0.81) | 0.645 (SD=0.008) |
| gpt-4o_ft | 3.8 (SD=0.76) | 4.2 (SD=0.76) | 4.3 (SD=0.72) | 3.48 (SD=0.72) | 0.628 (SD=0.089) |

*Standardized results*

## 4. Discussion

The evaluation of plain language adaptation in the biomedical domain is challenging and the traditional quantitative evaluation metrics mostly do not perform well compared to the qualitative (human) evaluations. For example, readability scores such as FK grade level provide a standardized, objective measure of text complexity, however they do not capture the nuances of the language and needs of the target audience [8,9], which could be observed in the higher-than-expected reading level of the ground truth. We saw a similar behavior for SARI, which also preferred FT approaches. Differences for complex scores, such as BERTScore, between approaches were also minimal. Only the LLM based G-Eval evaluation performed similarly to qualitative evaluation.

It is interesting to note that the smaller `gpt-4o-mini` model outperformed the bigger `gpt-4o` and that only using a prompt template performed almost as good as using the two AI agent approach, which iteratively improves the adaptation. The FT approaches, which had a similar text complexity as the ground truth, did not perform well neither in the qualitative evaluation nor when using G-Eval.

Some of the limitations of the study: only two proprietary OpenAI models were used, for private healthcare data a locally deployed LLMs should be considered/used; small sample dataset from curated biomedical abstracts might differ from real world long complex medical texts written by physicians or other healthcare experts; benchmark contamination analysis was not performed which might be concern, since there are signs that newer LLMs are contaminated with public benchmark datasets [13]; the small sample human evaluations using Likert point scales may be subject to bias and inconsistency in ratings.

## 5. Conclusions

We looked at some classic readability score, such as FK grade level and SMOG index, focusing on the former since the latter is more appropriate on longer texts. Qualitative evaluations are still the gold standard for custom task such as evaluation plain language adaptations, however G-Eval, a LLM based quantitative showed promising results, ranking the approaches similarly as the qualitative metric and in the future we intent to explore such LLM based evaluation metrics further as well as broaden the study.

## Acknowledgements

Authors acknowledge support from the Slovenia Research Agency [N3-0307, GC-0001] and European Union under the Horizon Europe [101159018].